\documentclass{article}

\usepackage{arxiv}

\usepackage[utf8]{inputenc} 
\usepackage[T1]{fontenc}    
\usepackage{hyperref}       
\usepackage{url}            
\usepackage{booktabs}       
\usepackage{amsfonts}       
\usepackage{nicefrac}       
\usepackage{microtype}      
\usepackage{lipsum}		
\usepackage{graphicx}
\usepackage{natbib}
\usepackage{doi}
\usepackage{pdfpages}

\begin{document}
%

\title{Are Bigger Encoders Always Better in Vision Large Models?}
\author{Bozhou Li$^\dagger$, Hao Liang$^\dagger$, Zimo Meng$^\dagger$, Wentao Zhang\\
Peking University\\
}
\maketitle
\begin{abstract}
\begin{quote}
In recent years, multimodal large language models (MLLMs) have shown strong potential in real-world applications. They are developing rapidly due to their remarkable ability to comprehend multimodal information and their inherent powerful cognitive and reasoning capabilities. Among MLLMs, vision language models (VLM) stand out for their ability to understand vision information. However, the scaling trend of VLMs under the current mainstream paradigm has not been extensively studied. Whether we can achieve better performance by training even larger models is still unclear. To address this issue, we conducted experiments on the pretraining stage of MLLMs. We conduct our experiment using different encoder sizes and large language model (LLM) sizes. Our findings indicate that merely increasing the size of encoders does not necessarily enhance the performance of VLMs. Moreover, we analyzed the effects of LLM backbone parameter size and data quality on the pretraining outcomes. Additionally, we explored the differences in scaling laws between LLMs and VLMs.
\end{quote}
\end{abstract}

\def\customfootnotetext#1#2{{%
  \let\thefootnote\relax
  \footnotetext[#1]{#2}}}

\customfootnotetext{1}{\textsuperscript{$\dagger$}Equal Contribution}

\section{Introduction}
In recent years, the rapid development of large language models has revolutionized the field of natural language processing ~\cite{devlin2019bert,touvron2023llama2openfoundation,openai2024gpt4,bai2023qwenvl,du2022glamefficientscalinglanguage}. These models, equipped with an enormous number of parameters, have demonstrated outstanding performance in areas such as translation ~\cite{fan2020englishcentric}, question answering ~\cite{devlin2019bert, Raffel2019ExploringTL}, and text generation ~\cite{brown2020languagemodelsfewshotlearners,du2022glamefficientscalinglanguage}. Due to the powerful capabilities of LLMs, a natural idea is to combine them with multimodal research, leveraging the strong cognitive abilities of LLMs to process information from other modalities ~\cite{openai2024gpt4,chu2024qwen2audiotechnicalreport, Maaz2023VideoChatGPTTD}. 

Among MLLMs, VLMs achieve competitive performance in traditional multimodal tasks such as image classification~\cite{chen2024internvl}, image understanding~\cite{li2023blip2}, and image captioning~\cite{bai2023qwenvl}. Moreover, their excellent language understanding capabilities enable strong performance in text-rich tasks, such as vision question-answering~\cite{llava,llava1.5} and image-text retrieval~\cite{chen2024internvl}.

Currently, the dominant architecture for VLMs employs a large language model as the backbone. Mechanisms such as cross-attention~\cite{li2023blip2} or linear projectors~\cite{llava} are utilized to connect the vision encoder (e.g., ViT~\cite{dosovitskiy2021imageworth16x16words}) with the LLM backbone~\cite{zhang2024mmllms}. Larger visual encoders possess stronger encoding capabilities, allowing them to extract features more effectively. These features are then transformed by the projectors for the LLMs to understand. We refer to this model architecture as the connected vision paradigm.

Apart from VLMs, previous research on scaling laws~\cite{kaplan2020scaling,hoffmann2022training} adopts an empirical approach to study the relationship between model performance, parameter size, and the amount of training data. Under the assumption that the scaling trend follows a power-law relationship, researchers fit power-law formulas by training models of different sizes with varying amounts of data. Additionally, scaling laws have been extended to various scenarios, including graph neural networks~\cite{liu2024neural}, data mixing~\cite{ye2024data}, data pruning~\cite{Sorscher2022BeyondNS}, and fine-tuning LLMs~\cite{zhang2024scaling}. Although~\citet{aghajanyan2023scaling} investigated scaling laws in the context of multimodal scenarios, they neglected the connected vision paradigm~\cite{zhang2024mmllms,bai2023qwenvl}, leading to the following challenges:

\textbf{C1. Poor Effectiveness:} Since no scaling laws have been conducted, it is uncertain whether this paradigm can scale up to achieve better performance, resulting in suboptimal model performance.

\textbf{C2. Low Efficiency:} Without scaling laws, we do not know how much data or how many parameters to use. Typically, all available data are used for training VLMs, resulting in a waste of data and computational resources.

To address these issues, we conduct scaling law experiments following~\cite {kaplan2020scaling}. We select the renowned LLaVA1.5~\cite{llava1.5} model as the backbone of our study. To investigate the scaling laws, we utilize models with 7 billion and 13 billion parameters. For the dataset, we choose image-text pairs from CC12M~\cite{changpinyo2021conceptual} and Laion400M~\cite{Schuhmann2021LAION400MOD}, known for their high-quality pairs. We use data sizes from 1 million to 10 million image-text pairs for our scaling law experiments.


The core contributions of this paper are summarized as follows:
\begin{itemize}
    \item \textbf{New Perspective} To the best of our knowledge, we are the first to conduct scaling law studies for the connect vision paradigm. Additionally, we are the first to analyze the limitations of this VLM paradigm.

    \item \textbf{New Observation} We conducted experiments on the pretraining phase of MLLMs using the connect vision paradigm. From the experimental results, we derived a crucial conclusion: simply utilizing a ViT trained using CLIP with more parameters and better performance does not enhance the performance of MLLMs. This suggests that exploring alternative methods is necessary to improve the performance of MLLMs. We also indicate that this issue is highly likely not caused by limitations in the scaling ability of ViT.
\end{itemize}
\section{Background and Related Work}
\subsection{Multimodal Large Language Model}

In recent years, benefiting from the development of model architectures based on transformers ~\cite{Vaswani2017AttentionIA} and the increase in computing resources and training data, LLMs ~\cite{devlin2019bert,touvron2023llama2openfoundation,openai2024gpt4,bai2023qwenvl,du2022glamefficientscalinglanguage} have emerged. These models possess massive parameters, requiring extensive computational resources to train on large datasets for a prolonged period. The sheer scale of these models, often exceeding billions of parameters, has enabled them to capture subtle nuances of language and learn complex patterns from vast amounts of text data. LLMs have not only demonstrated outstanding performance in traditional NLP tasks but have also exhibited emergent capabilities such as reasoning, code generation, and solving mathematical problems ~\cite{wei2022emergentabilitieslargelanguage}.

With the fast-paced advancement of LLMs, a natural question arises: how can we enhance the capabilities of LLMs to enable them to understand information from other modalities? Some earlier MLLMs, such as BEiT-3, attempted to train a multimodal model from scratch using data from multiple modalities ~\cite{Wang2022ImageAA}. However, there is now a growing preference for combining pre-trained LLMs and modality encoders using specific mechanisms to construct multimodal LLMs. The methods combining pre-trained LLMs and modality encoders can be classified into four categories ~\cite{wadekar2024evolutionmultimodalmodelarchitectures}. The first and second categories employ cross-modal attention mechanisms ~\cite{alayrac2022flamingo} and custom layers ~\cite{cho2021unifyingvisionandlanguagetaskstext}, respectively, to integrate information from other modalities into the internal representation of the LLM. These two types of methods are commonly referred to as deep fusion methods. The third category uses different modules to align information from other modalities with text information ~\cite{llava,li2023blip2,bai2023qwenvl}. Researchers exploring the fourth category attempt to unify the encoding of different modalities and expand the vocabulary of LLMs ~\cite{zhan2024anygpt}. These two methods are commonly referred to as early fusion methods. Due to their ease of training, high computational efficiency, and ability to construct any-to-any MLLMs, early fusion methods, especially the third category, which is also the focus of this article, are gradually becoming mainstream.

In general, the fusion modules used in the third category of methods can be classified into three types. The first type is transformer-based abstractors. These methods allow for adjusting the number of visual tokens to balance model performance and efficiency ~\cite{li2023blip2,bai2023qwenvl}. However, they might lose locality inductive bias and require longer training times and more training data. The second type involves using MLPs (Multi-Layer Perceptrons) ~\cite{llava}. This approach is lightweight, easy to train, and capable of capturing local information. However, it can only generate fixed-length visual tokens. Furthermore, some researchers have explored the use of custom layers to combine the advantages of both approaches ~\cite{cha2024honeybee}.

The training process of  MLLMs can be divided into two stages: multimodal pretraining and multimodal instruction fine-tuning. During the multimodal pretraining (MM PT) stage, the input and output projectors are trained to align different modalities by optimizing predefined objectives. Typically, X-text pairs are used in this stage, where X represents data from other modalities. In the multimodal instruction fine-tuning (MM IT) stage, the pre-trained multimodal LLM is fine-tuned using a dataset formatted in an instructional manner. This fine-tuning aims to enhance the model's generalization ability on unseen tasks. In this article, we conducted experiments on the MM PT stage.

\subsection{Scaling Law}

In the process of deep learning development, there has been a tendency to train larger models to achieve better performance. A natural question arises: as the number of model parameters and the amount of training data increase, how does model performance change? This has led to research on scaling laws. Hestness et al. ~\cite{Hestness2017DeepLS} were the pioneers in using empirical methods to study the scaling law of deep neural networks. Kaplan et al. ~\cite{kaplan2020scaling} and Hoffmann et al. ~\cite{hoffmann2022training} conducted extensive experiments on transformer-based autoregressive models at different scales. They investigated the relationship between model loss and the number of model parameters and training data.

In situations where the model architecture is known, we can approximate the training cost based on the number of model parameters and the amount of training data (e.g., $C \approx 6ND$ for the transformer-based model, where N represents the number of model parameters, D represents the amount of training data, and C represents the computational costs). By combining this estimation with scaling laws, we can determine the optimal allocation of training data and model parameters within limited computational resources ~\cite{hoffmann2022training}. This approach is particularly useful for LLMs since training them requires a significant amount of computational resources. Even if we embrace the philosophy underlying LLAMA ~\cite{touvron2023llama2openfoundation}, which entails setting aside computational expenses during the training phase and maximizing data usage to augment model performance at inference time, we consequently face an additional consideration: using larger models inherently translates to increased computational costs during inference. Therefore, even under conditions of ample data availability, a critical question arises: is the performance boost from expanding model parameter counts justified by the resultant higher computational overhead at inference? Under this assumption, scaling laws are still necessary.

After the initial scaling law was proposed, researchers studied the scaling law in different scenarios, such as graph neural networks ~\cite{liu2024neural}, data mixing ~\cite{ye2024data}, data pruning ~\cite{Sorscher2022BeyondNS}, and fine-tuning of large language models ~\cite{zhang2024scaling}. Some of them made certain adjustments to the form of the formula.

When it comes to MLLMs, Aghajanyan et al. ~\cite{aghajanyan2023scaling} explored the relationship between the loss and the dataset size and the parameter size over seven modalities, including text, image, image-text, speech, speech-text, code, and molecules. They also examined the scaling law when models are pre-trained on two modalities simultaneously and modified the form of the formula. However, the training paradigm they used in their study differs from the connected vision paradigm, the mainstream methods preferred today. They trained LLMs and modality encoders from scratch, similar to the approach chosen by BEiT-3 ~\cite{Wang2022ImageAA}, to obtain their scaling law. This discrepancy has sparked our research focus. To the best of our knowledge, the scaling law of multimodal large models under the connected vision paradigm is still under-researched.

\section{Methods}
\subsection{Dataset}

We utilized the CC12M dataset ~\cite{changpinyo2021conceptual} in our experiments, a comprehensive and expansive multimodal collection designed to foster advancements in artificial intelligence, particularly focusing on the intersection of vision and language. From the CC12M dataset, we strategically extracted subsets of data ranging from 1 million to 10 million images, incrementing by 1 million each time, to serve as our training datasets. This approach was chosen to systematically investigate the impact of varying training data sizes on the model's performance and learning outcomes. In our experiments involving different variants and scales of the Vision Transformer (ViT) trained by CLIP~\cite{Radford2021LearningTV}, we maintained the use of identical datasets for each training session. This decision was made to eliminate the influence of random variables that could arise from using different data splits. By controlling for the input data, we aimed to isolate and observe the effects of model architecture and size on the experimental results, providing a clearer understanding of how these factors influence performance.

To validate the generality of our findings, we also conducted a set of experiments using the LAION-400M dataset ~\cite{Schuhmann2021LAION400MOD}. Similarly, we constructed training datasets ranging from 1M to 10M by randomly sampling from the LAION-400M dataset.

For each image-text pair in the dataset, we randomly selected a text prompt from the LLaVA paper's text prompts ~\cite{llava} and randomly arranged its order with the image. We used the caption of the image as the ground truth. For each size of the training dataset, we extracted 25,600 of them as validation sets.

\subsection{Model Choice}
In our quest to elucidate how different sizes of the ViT influence the performance of multimodal large language models, we deemed it crucial to employ ViTs of various sizes. To ensure that extraneous factors such as disparities in training data distribution or variations in training recipes did not skew our findings, we meticulously opted for a consistent series of ViT models throughout our experiments.

Our choice fell upon the ViT models ~\cite{ilharco_gabriel_2021_5143773} trained using CLIP on the extensive Laion2B ~\cite{Schuhmann2022LAION5BAO} dataset, renowned for its comprehensive coverage and quality, thereby providing a robust baseline for our investigations. The specific model sizes we selected are shown in Table \ref{tab:example_vit}.

\begin{table}[h]
\centering
\caption{Parameter sizes of various CLIP models and their ViT modules from the LAION dataset.}
\begin{tabular}{ccc}  
\toprule  
\textbf{Model Name} & \textbf{Parameter Size of CLIP Model} & \textbf{Parameter Size of ViT Modules} \\
\midrule  
laion/CLIP-ViT-B-16-laion2B-s34B-b88K & 150M & 86M \\
laion/CLIP-ViT-L-14-laion2B-s32B-b82K & 428M & 304M \\
laion/CLIP-ViT-H-14-laion2B-s32B-b79K & 986M & 632M \\
laion/CLIP-ViT-g-14-laion2B-s34B-b88K & 1.37B & 1.01B \\
\bottomrule  
\end{tabular}
\label{tab:example_vit}
\end{table}

For the LLM backbone, the Vicuna series ~\cite{vicuna2023} was selected. This model series was cultivated by fine-tuning the LLaMA2 ~\cite{touvron2023llama2openfoundation} foundational model on a diverse set of user-shared dialogues sourced from ShareGPT ~\cite{ShareGPT}. To ensure the reliability of our conclusions, we conducted training on both the 7B model and the 13B model from the Vicuna series.
\subsection{Training Process}
In our overarching approach, we adopted the training process from the MM PT stage of LLaVA v1.5 ~\cite{llava1.5}. Specifically, a two-layer Multi-Layer Perceptron (MLP) served as the linear projector within our framework, utilizing the Gaussian Error Linear Units (GELU) ~\cite{Hendrycks2016GaussianEL} as the activation function to enhance non-linearity. Throughout the training phase, we froze the parameters of the ViT and LLM backbone and only updated the Linear Projector. 

We conducted the MM PT stage of the model using next token prediction as the training task and cross-entropy as the loss function. During the training process, we only masked the textual data in the ground truth and did not mask the image tokens.

Across a spectrum of dataset sizes, ranging from smaller to larger volumes, and varying the scale of ViT parameters and the size of the Vicuna model parameters, we trained the linear projector from scratch with a cosine scheduler. All experiments were conducted on 8*A100/A800 NVIDIA GPU machines.
\section{Experiments Results}
\begin{figure}[htbp]
\centering
\begin{minipage}{0.45\textwidth}
  \centering
  \includegraphics[width=\linewidth]{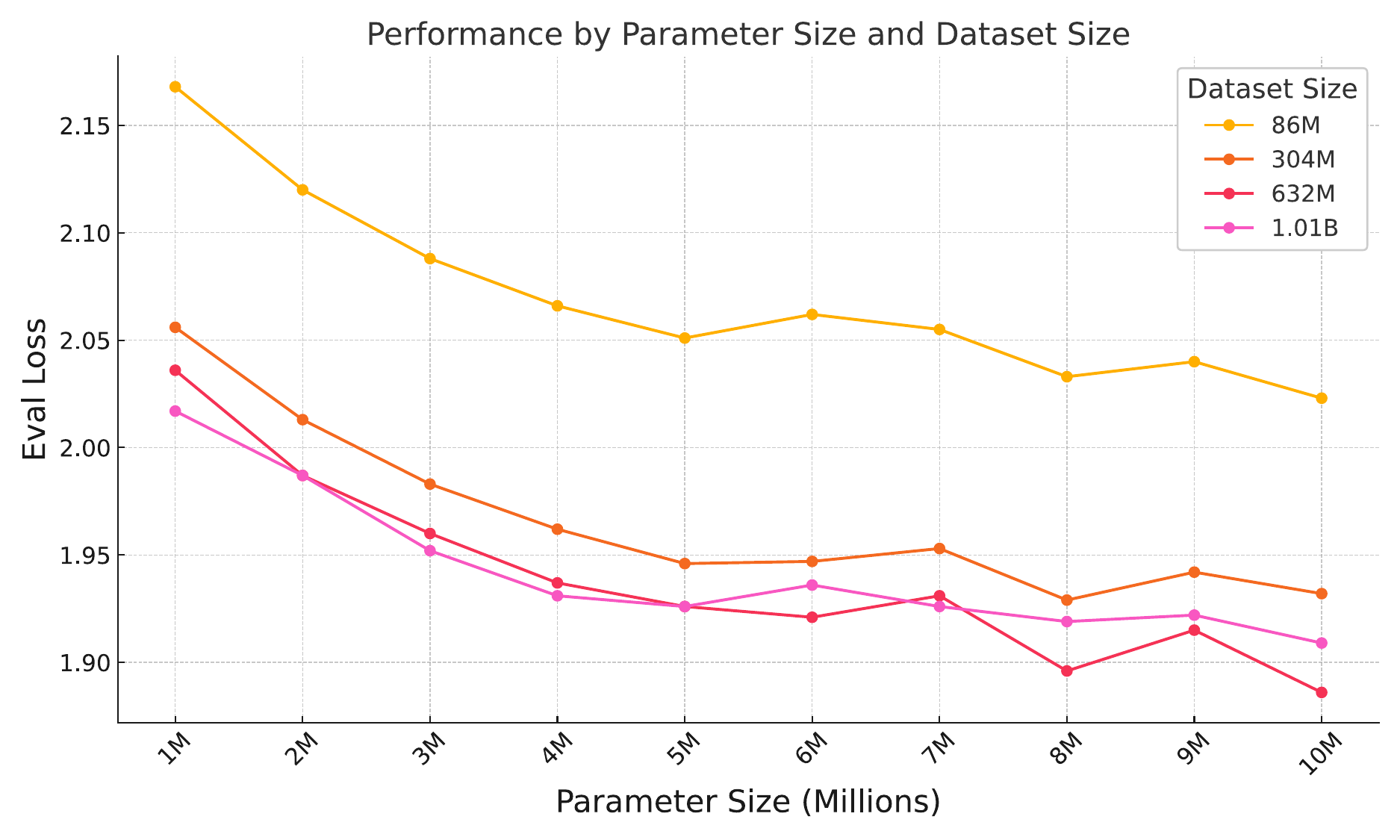} 
  \caption{Using Vicuna-7B as LLM backbone, trained on data sampled from CC12M}
  \label{fig:Vicuna-7B}
\end{minipage}
\hfill
\begin{minipage}{0.45\textwidth}
  \centering
  \includegraphics[width=\linewidth]{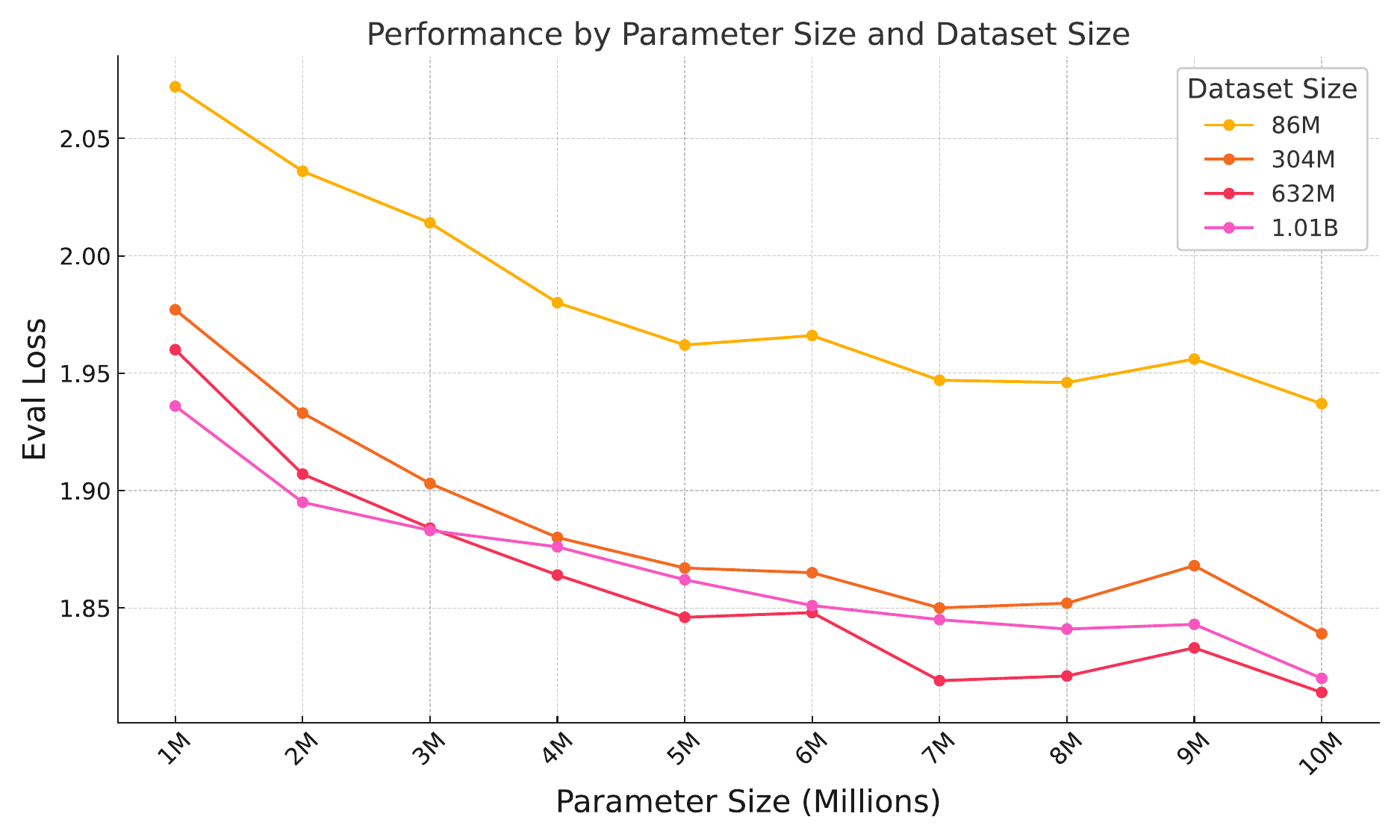} 
  \caption{Using Vicuna-13B as LLM backbone, trained on data sampled from CC12M}
  \label{fig:Vicuna-13B}
\end{minipage}
\hfill
\begin{minipage}{0.45\textwidth}
  \centering
  \includegraphics[width=\linewidth]{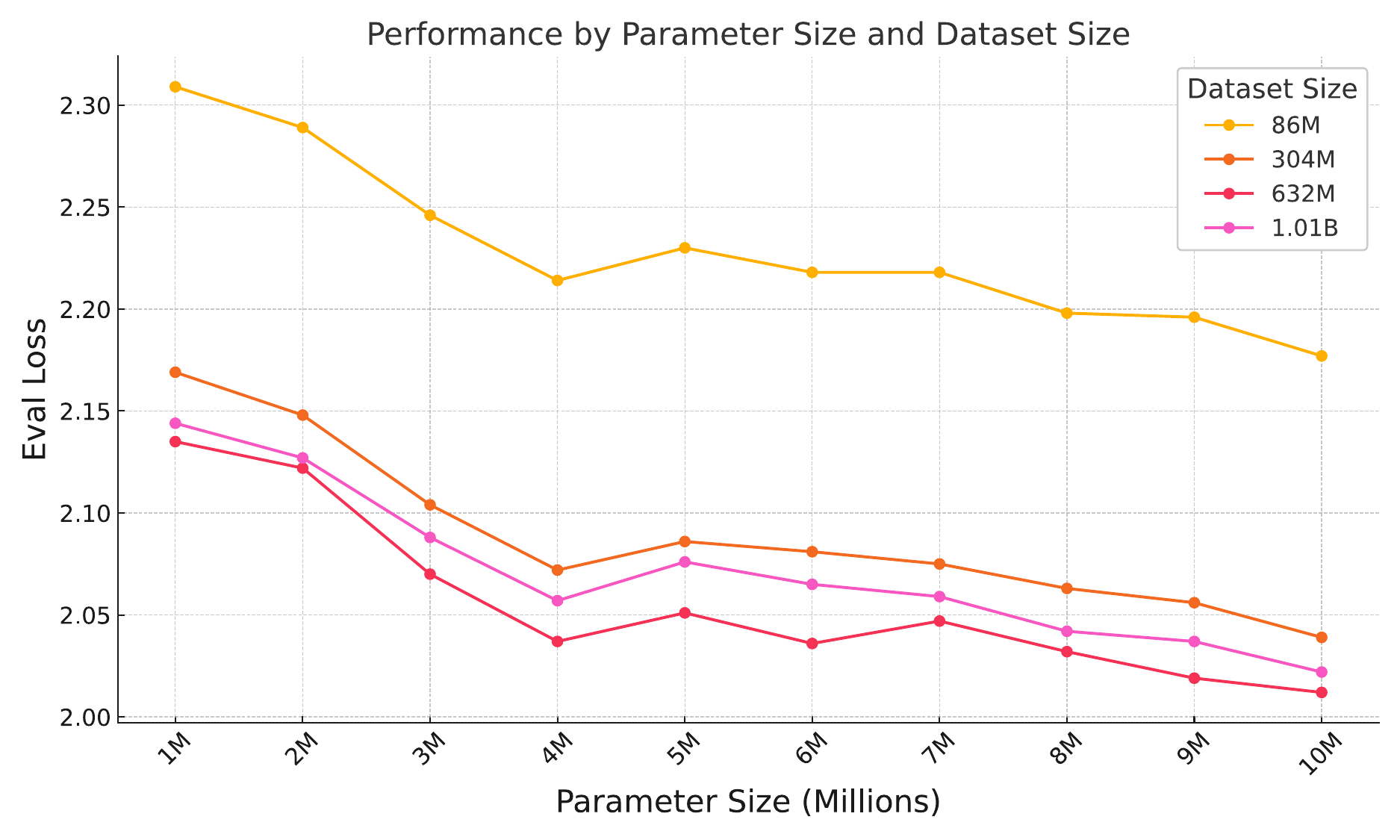} 
  \caption{Using Vicuna-7B as LLM backbone, trained on data sampled from LAION-400M}
  \label{fig:Vicuna-7B-laion400m}
\end{minipage}
\end{figure}

Tables  \ref{tab:vicuna-7B} and  \ref{tab:vicuna-13B} exhibit the concluding evaluation loss values obtained when employing various dataset sizes and ViT model sizes and training datasets built from CC12M. To render these outcomes more comprehensible at a glance, we have also produced Figures \ref{fig:Vicuna-7B}  and  \ref{fig:Vicuna-13B}. These visual aids aim to depict the relationship between model size, dataset size, and training loss, making the trends and patterns easier to discern. Table \ref{tab:vicuna-7B-laion400m} and Figure \ref{fig:Vicuna-7B-laion400m} show the results with datasets built from LAION-400M.

Drawing insights from our experimental findings, several noteworthy observations emerge:
\begin{itemize}
    \item \textbf{Increasing Data Quantity Improves Model Performance}
    
    It is observed that for smaller training datasets, augmenting the data quantity effectively reduces the evaluation loss. This aligns with the learning theory that more examples lead to better model performance.
    \item \textbf{Larger LLM Backbone Enhances Model Performance} 
    
    Comparing the training outcomes on the Vicuna-7B and Vicuna-13B models reveals that larger models consistently present lower evaluation loss than their smaller versions. This implies that larger language models possess a heightened capability to understand and interpret vision tokens.
    \item \textbf{The Importance of High-Quality Data}
    
    The evaluation loss obtained from training on the CC12M dataset is lower than the evaluation loss from training on the LAION-400M dataset. The ViT models used in the experiment were trained on the LAION-2B dataset. Intuitively, its distribution should be closer to LAION-400M, and training multimodal LLMs with LAION-400M would yield better results. However, experimental results show that using CC12M is more effective. This is likely due to the quality of the datasets. Compared to LAION-400M, the CC12M dataset has higher quality, with a higher degree of matching between image and text. This result emphasizes the importance of high-quality datasets when training MLLMs.
    \item \textbf{Larger LLM Backbone Requires Less Training Data} 
    
    Notably, when leveraging the Vicuna-13B model, the evaluation loss plateaus around the 7M mark in training data size, signifying a diminishing return on additional data. Conversely, the Vicuna-7B model continues to benefit from an increased data volume, illustrating a more gradual learning curve. This indicates that smaller models might require more substantial data increments to observe subsequent improvements, while larger models can exhibit more pronounced progress with less data.
    \item \textbf{Improved ViT Performance Doesn't Guarantee Better Results}
    
    A critical finding from our study reveals that merely amplifying the parameter scale of ViT does not necessarily translate into enhanced model performance. Although vision transformers exhibit a relatively lesser scaling capacity compared with text transformers ~\cite{zhai2022scalingvisiontransformers}, the parameter scales selected in our experiments are significantly removed from the upper limits of ViT's scaling capabilities.

The 632M ViT trained using CLIP achieves a zero-shot accuracy of 78.0\% on the ImageNet-1K dataset ~\cite{imagenet15russakovsky} while increasing the model parameter size to 1.01B and training on a larger amount of data results in a zero-shot accuracy leap to 78.4\% ~\cite{ilharco_gabriel_2021_5143773}. Moreover, through architectural advancements, researchers have demonstrated ViT models boasting accuracies above 90\%, with parameter counts around 20B ~\cite{zhai2022scalingvisiontransformers, Dehghani2023ScalingVT}. This underscores that within our experimental parameters, ViT's scaling potential has not been saturated, with enhanced performance still achievable through augmented model dimensions.

However, within our specific scenario, increasing the ViT's size does not yield superior results and, in certain instances, may lead to performance degradation. When training with the CC12M dataset, the validation loss of multimodal LLMs constructed with the 1.01B CLIP model's ViT is lower than that of models using the 632M model only when the training data volume is large. However, when training with data from the LAION-400M dataset, the validation loss of models using the 1.01B model consistently remains lower than that of models using the 632M model. This suggests that additional factors are at play, hindering the alignment and effectiveness between ViT and LLM in MM PT stage, necessitating deeper investigation.

\end{itemize}

\begin{table}[ht]
\centering
\caption{Evaluation Loss Using Vicuna-7B as LLM Backbone, Trained on Data Sampled from CC12M}
\begin{tabular}{@{}lcccccccccc@{}}  
\toprule  
\textbf{Parameter Size} & \textbf{1M} & \textbf{2M} & \textbf{3M} & \textbf{4M} & \textbf{5M} & \textbf{6M} & \textbf{7M} & \textbf{8M} & \textbf{9M} & \textbf{10M} \\
\midrule  
86M  & 2.168 & 2.120 & 2.088 & 2.066 & 2.051 & 2.062 & 2.055 & 2.033 & 2.040 & 2.023 \\
304M & 2.056 & 2.013 & 1.983 & 1.962 & 1.946 & 1.947 & 1.953 & 1.929 & 1.942 & 1.932 \\
632M & 2.036 & \textbf{1.987} & 1.960 & 1.937 & \textbf{1.926} & \textbf{1.921} & 1.931 & \textbf{1.896} & \textbf{1.915} & \textbf{1.886} \\
1.01B & \textbf{2.017} & \textbf{1.987} & \textbf{1.952} & \textbf{1.931} & \textbf{1.926} & 1.936 & \textbf{1.926} & 1.919 & 1.922 & 1.909 \\
\bottomrule  
\end{tabular}
\label{tab:vicuna-7B}
\end{table}

\begin{table}[ht]
\centering
\caption{Evaluation Loss Using Vicuna-13B as LLM Backbone, Trained on Data Sampled from CC12M}
\begin{tabular}{@{}lcccccccccc@{}}  
\toprule  
\textbf{Parameter Size} & \textbf{1M} & \textbf{2M} & \textbf{3M} & \textbf{4M} & \textbf{5M} & \textbf{6M} & \textbf{7M} & \textbf{8M} & \textbf{9M} & \textbf{10M} \\
\midrule  
86M & 2.072 & 2.036 & 2.014 & 1.980 & 1.962 & 1.966 & 1.947 & 1.946 & 1.956 & 1.937 \\ 
304M & 1.977 & 1.933 & 1.903 & 1.880 & 1.867 & 1.865 & 1.850 & 1.852 & 1.868 & 1.839 \\ 
632M & 1.960 & 1.907 & 1.884 & \textbf{1.864} & \textbf{1.846} & \textbf{1.848} & \textbf{1.819} & \textbf{1.821} & \textbf{1.833} & \textbf{1.814} \\ 
1.01B & \textbf{1.936} & \textbf{1.895} & \textbf{1.883} & 1.876 & 1.862 & 1.851 & 1.845 & 1.841 & 1.843 & 1.820 \\ 
\bottomrule  
\end{tabular}
\label{tab:vicuna-13B}
\end{table}

\begin{table}[ht]
\centering
\caption{Evaluation Loss Using Vicuna-7B as LLM Backbone, Trained on Data Sampled from LAION-400M}
\begin{tabular}{@{}lcccccccccc@{}}  
\toprule  
\textbf{Parameter Size} & \textbf{1M} & \textbf{2M} & \textbf{3M} & \textbf{4M} & \textbf{5M} & \textbf{6M} & \textbf{7M} & \textbf{8M} & \textbf{9M} & \textbf{10M} \\ 
\midrule  
86M & 2.309 & 2.289 & 2.246 & 2.214 & 2.230 & 2.218 & 2.218 & 2.198 & 2.196 & 2.177 \\ 
304M & 2.169 & 2.148 & 2.104 & 2.072 & 2.086 & 2.081 & 2.075 & 2.063 & 2.056 & 2.039 \\ 
632M & \textbf{2.135} & \textbf{2.122} & \textbf{2.070} & \textbf{2.037} & \textbf{2.051} & \textbf{2.036} & \textbf{2.047} & \textbf{2.032} & \textbf{2.019} & \textbf{2.012} \\ 
1.01B & 2.144 & 2.127 & 2.088 & 2.057 & 2.076 & 2.065 & 2.059 & 2.042 & 2.037 & 2.022 \\ 
\bottomrule  
\end{tabular}
\label{tab:vicuna-7B-laion400m}
\end{table}

\section{Discussion \& Conclusion}\label{sec:5}
Our experimental outcomes underscore a critical observation: within the LLaVA ~\cite{llava} framework, the simplistic augmentation of ViT's parameter scale does not correspondingly elevate the performance of MLLMs. This finding necessitates a strategic shift in our approaches to multimodal model design and optimization.
\begin{itemize}
    \item \textbf{Strengthen Data Alignment Across Modalities With Data-centric Approaches}
        
The experimental insights suggest the importance of improving the quality of training data for the alignment of different modalities. In this pursuit, data-centric approaches appear to hold significant promise. By focusing on enhancing data quality, variety, and relevance, we can potentially achieve better alignment between visual and textual modalities with higher efficiency, potentially leading to superior performance in multimodal applications ~\cite{bai2024surveymultimodallargelanguage}.
        
    \item \textbf{Investigate The Differences Between Various Alignment Methods}
    
    Deep Dive into Underlying Performance Limitations: A comprehensive investigation into why ViT performance enhancements do not translate into superior multimodal model performance is warranted. This exploration should encompass the dynamics of data utilization, model architecture, and training methodologies. When comparing the training approaches of CLIP ~\cite{Radford2021LearningTV} and LLaVA ~\cite{llava} we can observe certain differences in alignment strategies. The popular CLIP model utilizes a decoder-only LLM as the text encoder, such as GPT-2 ~\cite{radford2019language}. In CLIP, the CLS token of the image processed by ViT is aligned with the EOS token of the text processed by GPT-2. In contrast, the processing approach of LLaVA is more akin to mapping the image tokens processed by ViT to the embedding space of a LLM using MLP layers. This differs from aligning image tokens with text tokens processed by GPT-2, as done in CLIP. The impact of these alignment method differences on model performance is worth exploring.

    \item \textbf{Exploration of Multimodal Information Fusion}
    \begin{itemize}
        \item \textbf{Architectural Innovations for Multimodal Integration:} There is a need to explore novel architectures that facilitate more seamless integration of multimodal information ~\cite{cha2024honeybee,li2023blip2}. This could involve designing frameworks that better enable the exchange and processing of visual, textual, and other modal data, potentially leading to enhanced overall performance.
        \item \textbf{Expansion of Vocabulary:} Considering the inclusion of techniques like AnyGPT~\cite{zhan2024anygpt} to extend the vocabulary can prove beneficial in enriching the model's understanding and processing capabilities across different modalities. This could potentially enable the model to capture and utilize a broader spectrum of information, thereby enhancing its performance.
    \end{itemize}
\end{itemize}

The observed phenomenon of larger LLMs requiring smaller datasets for alignment with ViT suggests a closer alignment between the semantic spaces of vision and text. This insight not only enriches our understanding of multimodal model dynamics but also opens avenues for optimizing data requirements and training strategies for MLLMs. Future research should delve deeper into the mechanisms behind these observations, aiming to leverage the inherent semantic alignment capabilities of larger LLMs for enhanced performance and efficiency in multimodal applications.

\section{Limitations and Future Work}
Despite our study encompassing an extensive series of one hundred and twenty experiments across various parameter combinations, the investigation into the scaling laws of MLLMs during the MM PT stage remains a vast and complex field. Our current exploration, while insightful, is not exhaustive in addressing the comprehensive nature of this topic.

Compared to the scaling law of LLMs ~\cite{kaplan2020scaling,hoffmann2022training}, the scaling law of MLLMs exhibits more complex characteristics.
\begin{itemize}
    \item \textbf{More Influencing Factors}
    
    More factors influence the performance of MLLMs, particularly when using pre-trained modality encoders and an LLM backbone to build MLLMs, the paradigm that is currently widely used. The parameter size of modality encoders, the amount of data used for pretraining the modality encoders and LLM backbone, the quality of data during the MM PT stage, and the distribution discrepancy between the data used for MM PT and the data used for pretraining the modality encoders and LLM backbone can potentially influence model performance. These factors may also interact with each other in more complex ways. In addition, different modalities, fusion methods for integrating multimodal information, and architectures of modality encoders may exhibit different forms of scaling law.
    \item \textbf{Different Estimates of Computational Costs}
    
    Under the current training framework, it is not appropriate to solely rely on $C\approx 6ND$  for estimating computational costs. Due to the utilization of the backpropagation algorithm to update the model parameters, even if we only update the parameters of the MLP layer, which is quite lightweight, we still need to compute gradients for a significant portion of the LLM parameters. In addition, the modality encoder only needs to process data from other modalities, while the LLM also needs to handle text tokens. Taking LLaVA as an example, the compute budget for the MM PT stage of LLaVA can be estimated as:
    \begin{equation}
        C \approx 2N_{ViT}D_{img}+6N_{LLM}(D_{img}+D_{txt})
    \end{equation}
    where $N_{LLM}$ and $N_{ViT}$ represent the number of parameters of LLM and ViT, and $D_{img}$ and $D_{txt}$ represent the number of image tokens and text tokens, respectively. The MLP layer is overlooked due to its relatively small number of parameters. It is worth noting that the value of $D_{img}$ is dependent on the architecture of ViT, while the ratio between $D_{txt}$ and $D_{img}$ is also dependent on the source of the image-text pairs. For other fusion methods and modality encoder architectures, this formula may have different forms. This introduces new challenges in making trade-offs between computational costs and model performance.
\end{itemize}

To firmly substantiate the conclusions drawn in this paper, a more extensive empirical foundation is required. This necessitates conducting additional experiments on a broader range of LLM families, encompassing diverse architectures and capabilities. Moreover, the exploration should be extended to incorporate a wider variety of datasets—both in terms of content and scale—to ensure a comprehensive understanding of how multimodal models behave under different conditions.

Beyond empirical evidence, the surprising outcomes observed in our experiments demand a theoretical underpinning that goes beyond mere data analysis. The unexpected findings, such as the non-intuitive scaling behaviors of multimodal models or the peculiar interactions between vision and language semantic spaces, call for a deeper level of understanding that can only be achieved through rigorous theoretical analysis.

The prohibitive cost associated with extensive experimental setups has constrained our investigation to a specific scenario: the performance of large models where both ViT and LLM components are frozen. This approach, while providing valuable insights into the behavior of multimodal models under resource constraints, deviates significantly from the methodologies employed by some mainstream multimodal models such as Qwen-vl ~\cite{bai2023qwenvl}.

Our current experimental endeavors have been confined to the LLaVA architecture, providing a focused exploration of scaling properties within this specific framework. However, the landscape of multimodal AI architectures extends far beyond LLaVA, encompassing a diverse array of designs that include transformer-based connector architectures ~\cite{li2023blip2} and VAE-based vision encoder architectures ~\cite{kim2021conditionalvariationalautoencoderadversarial}. Each of these architectural paradigms embodies unique characteristics and potential scaling behaviors that require systematic investigation.

Our experiments only considered MLLMs that can understand different modalities but can only generate textual information. We did not take into account more powerful yet complex MLLMs that can generate content in other modalities. The study of the scaling behavior of such models is more complex and requires additional computational resources.

\bibliographystyle{unsrtnat}
\bibliography{reference}
\end{document}